\documentclass[fleqn,10pt]{wlscirep}

\usepackage{xcolor,amsmath}

\usepackage{amsfonts}

\usepackage[mathcal]{eucal}

\title{\LARGE NeuroVLA: A Brain-inspired Embodied Intelligence for Fluid and Fast Reflexive Robotics Control}

\author[1,4]{Weiyu Guo}
\author[1,4]{He Zhang}
\author[1,4]{Pengteng Li}
\author[1,4]{Tiefu Cai}
\author[1,4]{Ziyang Chen}
\author[1,4]{Yandong Guo}
\author[4]{He Xiao}
\author[3*]{Yongkui Yang}
\author[1,2*]{Ying Sun}
\author[1,2*]{Hui Xiong}

\affil[1]{The Thrust of Artificial Intelligence, The Hong Kong University of Science and Technology (Guangzhou), Guangzhou, China.}
\affil[2]{The Department of Computer Science and Engineering, The Hong Kong University of Science and Technology, Hong Kong, China}
\affil[3]{Shenzhen Institutes of Advanced Technology, Chinese Academy of Sciences, Shenzhen, China}
\affil[4]{AI$^2$ Robotics, Shenzhen, China}
\affil[*]{The corresponding authors (yk.yang@siat.ac.cn, yings@hkust-gz.edu.cn, xionghui@ust.hk).}

\begin{abstract}
Vision-language-action (VLA) models inherit powerful semantic representations from large language models, but carrying over their design priorities can leave the demands of physical control under-specified. Stable manipulation requires high-frequency proprioceptive feedback and persistent motor state, in addition to instruction understanding. Here we introduce Neuromorphic Vision-Language-Action (NeuroVLA), a brain-inspired architecture that couples a cortical vision-language planner to recurrent cerebellar modulation and a stateful spiking spinal controller. These components translate semantic intent into actions while incorporating high-frequency proprioceptive and force feedback, maintaining short-term motor memory, and enabling local responses to unexpected contact. In simulated and physical manipulation tasks, the cerebellar module reduces mean commanded jerk by 75.6\%, and NeuroVLA achieves a 54.8\% collision-recovery rate compared with 0.0\% for the evaluated VLA baselines. The spinal controller implemented on an FPGA-based neuromorphic processor achieves 2.19 ms inference latency and 0.87 mJ per inference, corresponding to approximately 0.4 W during inference. These results motivate moving beyond inherited language-model design priorities toward embodied architectures that explicitly integrate semantic planning, high-frequency bodily feedback, motor memory, and local correction.

\end{abstract}

\hypersetup{
  pdftitle={NeuroVLA: A Brain-inspired Embodied Intelligence for Fluid and Fast Reflexive Robotics Control},
  pdfauthor={Weiyu Guo; He Zhang; Pengteng Li; Tiefu Cai; Ziyang Chen; Yandong Guo; He Xiao; Yongkui Yang; Ying Sun; Hui Xiong},
  pdfsubject={Neuromorphic vision-language-action and robotic control},
  pdfkeywords={NeuroVLA, neuromorphic, vision-language-action, robotics, reflexive control}
}

\begin{document}
\maketitle

\section{Introduction}

Vision-language-action (VLA) models extend pretrained language and vision-language models to robotic control, transferring semantic knowledge and instruction-following capabilities~\cite{zitkovich2023rt, kim2024openvla}. Their action interfaces range from token prediction to specialized continuous-action experts~\cite{wang_multimodal_2026, pi_0.5, pi_0.5_fast}. We argue that this productive continuity can also create \textit{architectural inertia}: design priorities established for language modeling continue to shape embodied systems, even when physical interaction calls for different feedback rates, internal states, and response pathways. Larger semantic models and more expressive action outputs do not, by themselves, specify how a controller should sense rapid bodily changes, retain motor context, or react locally to contact.

Embodied intelligence therefore requires explicit attention to the design of the motor system. High-frequency proprioception grounds control in the current physical state; persistent motor dynamics retain the recent history needed for coordinated execution; and local feedback pathways support rapid correction between semantic updates. These are engineering counterparts of biological sensorimotor principles, including the stateful, practiced execution often described as ``muscle memory.'' Their relevance is concrete: pouring liquid requires continuous stabilization as the load changes, shaking a flask requires tracking the phase and progress of repeated motions, and recovering from contact requires responding before a slow visual inference cycle completes.

\begin{figure*}[!t]
	\centering
	\includegraphics[width=17cm]{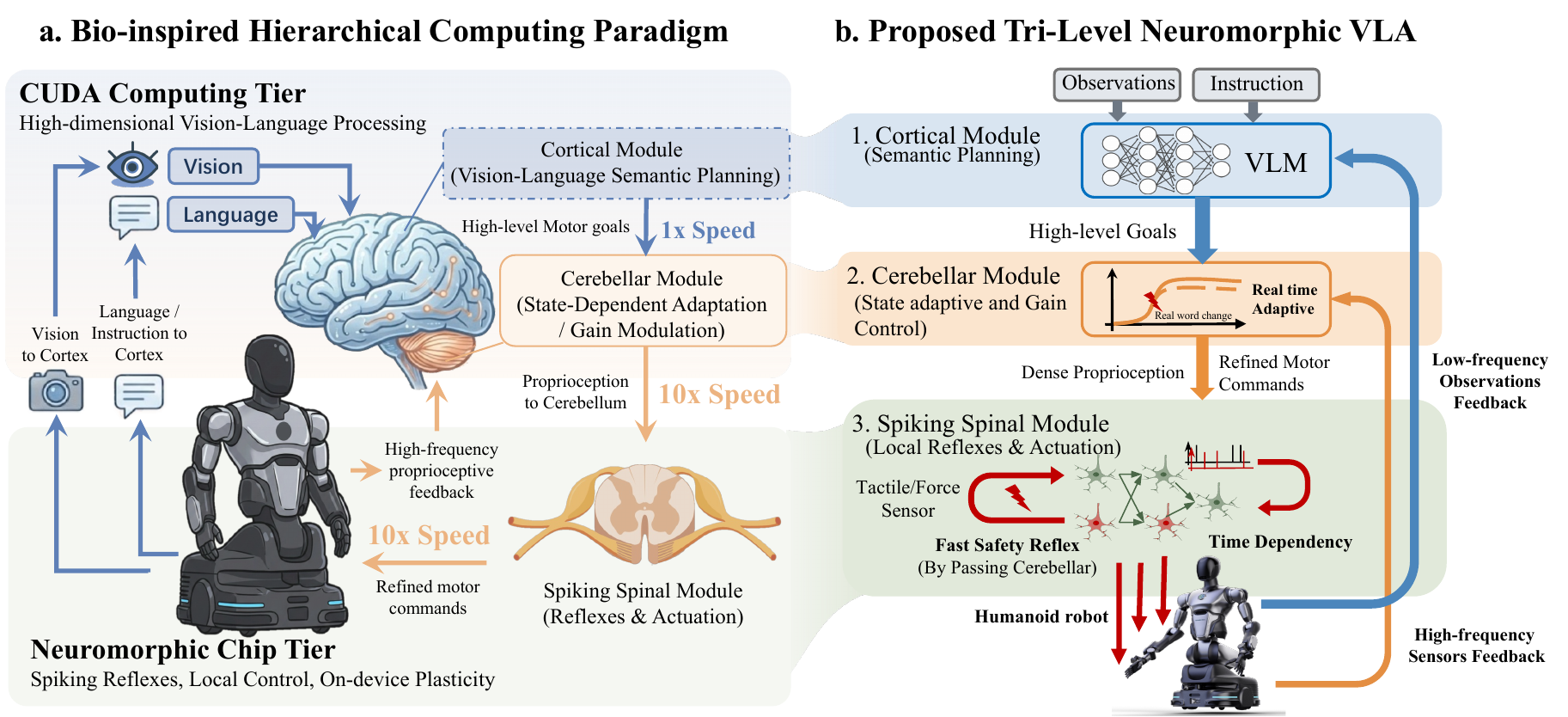}
	\caption{\textbf{Semantic planning coupled to feedback-driven motor control.}
\textbf{a,} Conceptual separation of high-dimensional semantic computation and local sensorimotor processing. The schematic illustrates the intended CUDA/neuromorphic partition; in the evaluated implementation, the cortical and cerebellar modules run on the CUDA tier, and hardware measurements concern the FPGA-based spinal controller.
\textbf{b,} Information flow in NeuroVLA. A cortical vision-language module generates compact intention tokens; recurrent cerebellar modulation combines these tokens with proprioceptive history; and a stateful spiking spinal module produces continuous actions. The force-feedback pathway supports local contact responses between cortical updates. }

	\label{fig:zhutu}
\end{figure*}

These tasks expose three related control bottlenecks. First, low-rate, vision-dominant feedback can miss rapid changes in joint state or contact force, allowing oscillation and trajectory drift to persist. Second, action decoders without persistent motor state must infer temporal progress from successive observations, making rhythmic and partially occluded tasks difficult. Third, routing every fine-grained adjustment through a large foundation model couples response time and computational cost to semantic inference. Efficient decoding and execution strategies address aspects of these problems~\cite{park2025saliency, zheng2025leveragingoslevelprimitivesrobotic, black2025realtime}; a complementary approach is to give local motor control access to bodily feedback and its own temporal state.

Biological motor systems motivate this separation of responsibilities. Cortical planning interacts with cerebellar prediction and correction, while spinal circuits integrate descending commands with local feedback~\cite{scott2004optimal, Ito2008InternalModels, bruel2024spinal}. We adopt these functional principles without assuming a one-to-one correspondence between engineered modules and anatomical subdivisions. In NeuroVLA, a cortical vision-language model supplies compact semantic intent through a Q-Former, a recurrent cerebellar module conditions that intent on proprioceptive history, and a stateful spiking spinal module produces continuous actions (Fig.~\ref{fig:zhutu}). High-frequency force feedback supports local contact responses, while recurrent and membrane states preserve short-term motor context. This architecture allows motor regulation to operate between semantic updates.

We test these mechanisms through attention analysis, proprioceptive control traces, temporal ablations, hardware measurements, and physical manipulation tasks. Cerebellar modulation reduces mean commanded jerk by 75.6\%; multi-step spiking dynamics improve task success over a single-step variant; and the FPGA-based spinal controller performs inference in 2.19 ms at 0.87 mJ per inference. On physical hardware, NeuroVLA outperforms the evaluated VLA baselines and achieves 54.8\% recovery in a collision stress test, compared with 0.0\% for those baselines. Together, these results identify feedback bandwidth, persistent motor state, and local correction as useful complements to high-level instruction understanding.

\section{Results}

\subsection{A Brain-inspired VLA with Event-driven Sparsity}
NeuroVLA separates semantic interpretation, feedback-based modulation, and action generation (Fig.~\ref{fig:zhutu}). The cortical module processes images and language instructions with a pretrained vision-language model. A Q-Former extracts compact, task-relevant intention tokens from its hidden representations~\cite{li2023blip}, providing the downstream controller with an action-oriented semantic interface.

The cerebellar module integrates a history of joint states and force/torque measurements through recurrent dynamics. Gated feature-wise linear modulation then adjusts the semantic latent according to the current physical context. This mechanism makes local stabilization and contact-aware correction possible without requiring a new cortical inference for every adjustment. Its functional inspiration is cerebellar prediction and feedback control~\cite{Ito2008InternalModels, Garrido2013Distributed}.

The spinal module converts the modulated latent into continuous actions using stateful spiking neurons. Membrane dynamics carry information across control steps, while selective spiking activity provides an event-driven execution substrate. In the implementation described in Methods, cortical and cerebellar computations run on the CUDA tier, and the spinal controller is validated on an FPGA-based neuromorphic processor. The following experiments examine semantic selection, feedback modulation, temporal integration, and physical performance in turn; architectural details are provided in Methods~\ref{methods}.

\begin{figure*}[!t]
	\centering
	\includegraphics[width=17cm]{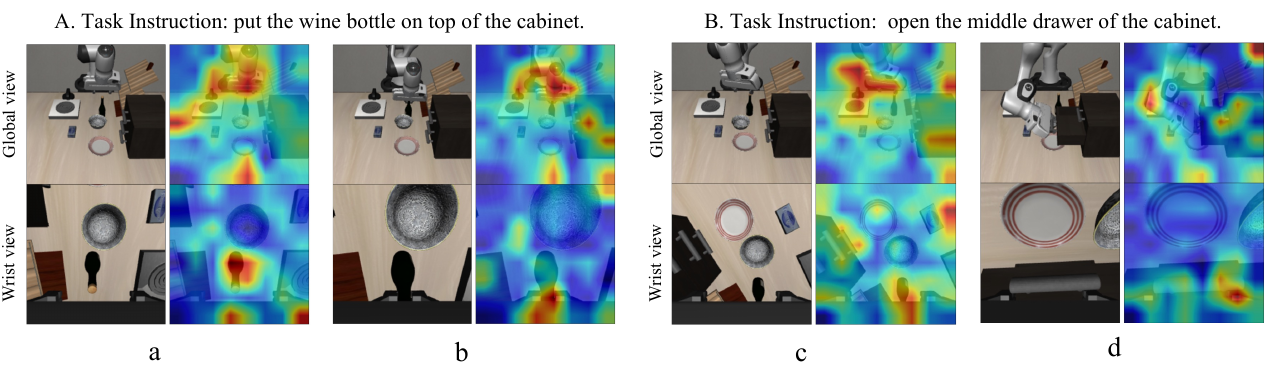}
    \vspace{-0.3cm}
	\caption{\textbf{Instruction-dependent spatial selection by the Q-Former.}
\textbf{a, b,} Attention during ``Put the wine bottle on the cabinet'': focus shifts from the manipulation target to the placement surface as execution progresses.
\textbf{c, d,} Under ``Open the middle drawer'' in the same scene, attention concentrates on the drawer handle despite the nearby bottle. These maps illustrate task-dependent selection of spatial features for downstream motor control.}

	\label{fig:atten}
\end{figure*}

\subsection{Task intent drives top-down attentional gating of spatial primitives}
\label{sec:qformer}
To examine how semantic intent selects spatial information, we visualized the Q-Former's cross-modal attention under different instructions in the same scene (Fig.~\ref{fig:atten}). For ``Put the wine bottle on top of the cabinet,'' attention first concentrated on the bottle and then shifted toward the placement surface as execution progressed. When the instruction changed to ``Open the middle drawer,'' attention instead concentrated on the drawer handle despite the bottle remaining visible.

These observations show instruction-dependent spatial selection, consistent with the Q-Former's role as an information bottleneck between visual-language features and motor control. They support task-relevant filtering before feedback-based motor modulation and action generation.

\begin{figure*}[!t]
	\centering
	\includegraphics[width=17cm]{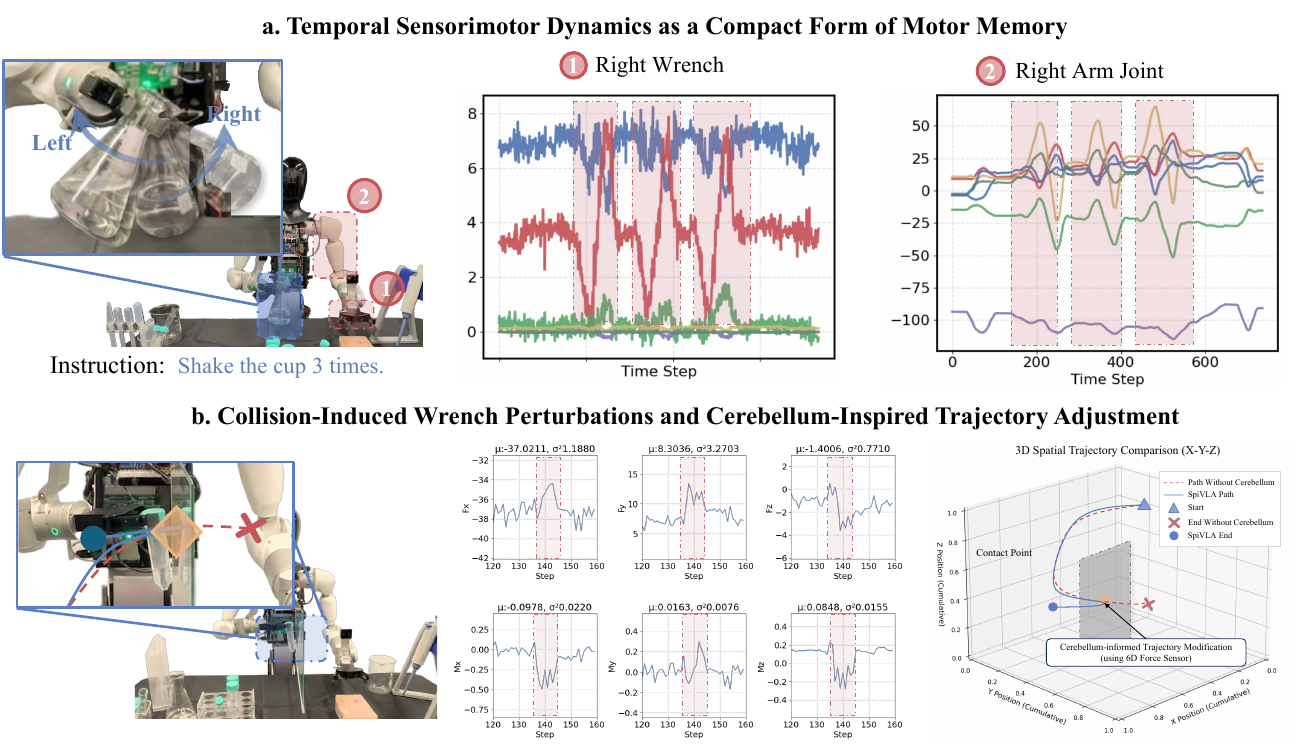}
	\caption{\textbf{Temporal coordination and force-aware trajectory correction.}
\textbf{a,} Wrench and joint-angle traces during ``Shake the cup'' show synchronized oscillatory cycles, providing a sensorimotor description of rhythmic execution.
\textbf{b,} Contact produces a wrench transient. The local feedback pathway adjusts the subsequent trajectory (blue solid line), whereas the baseline continues along the obstructed path (red dashed line). These examples assess temporal coordination and contact-aware correction in the engineered controller.}

	\label{fig:xianaoresult}
\end{figure*}

\subsection{Proprioceptive feedback stabilizes motion and supports local correction}

\begin{figure*}[!t]
	\centering
	\includegraphics[width=17cm]{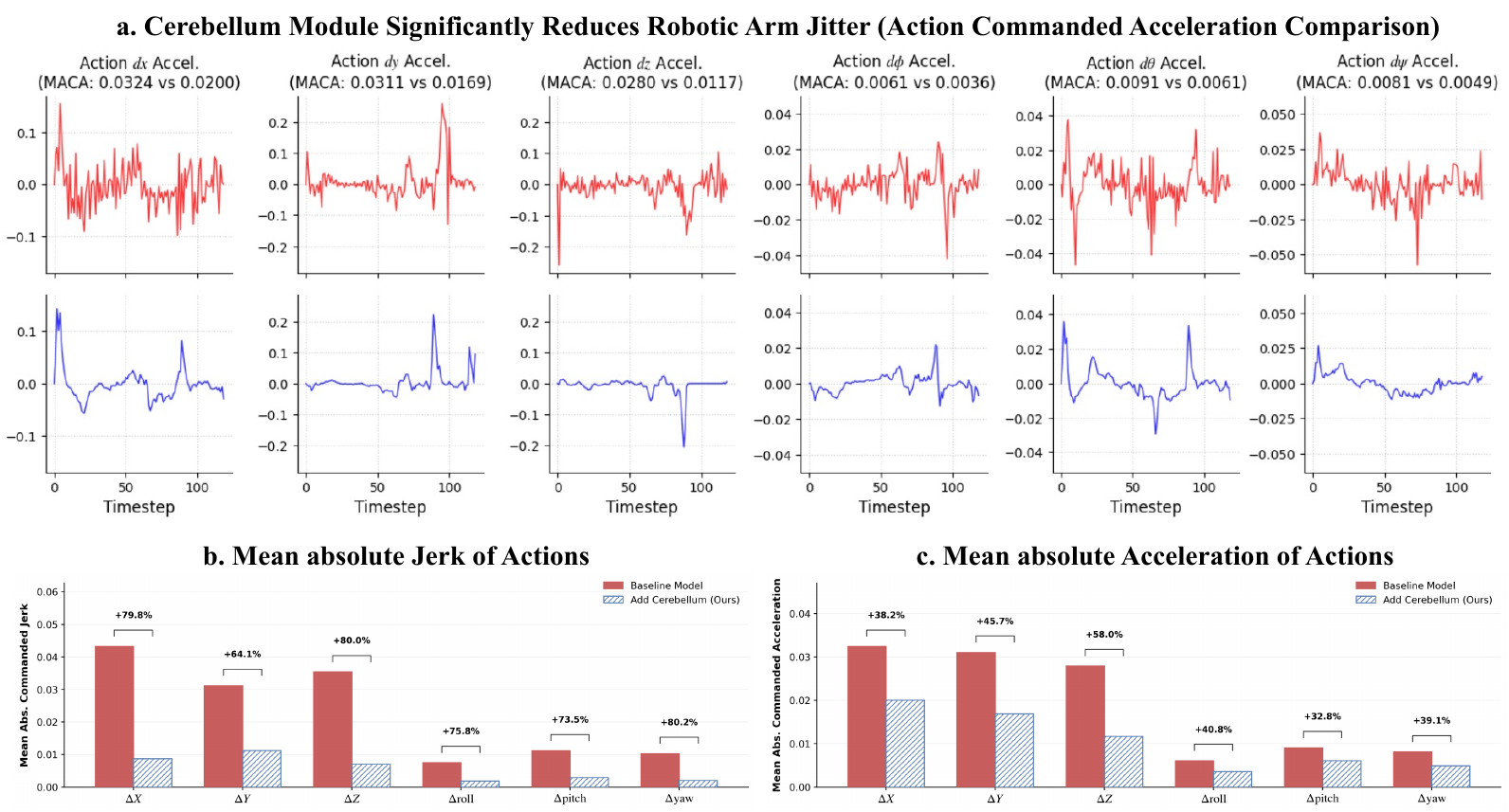}
	\caption{\textbf{Proprioceptive modulation reduces fluctuations in commanded motion.}
\textbf{a,} Commanded acceleration traces in translational and rotational dimensions with the baseline policy (red) and cerebellar modulation (blue).
\textbf{b, c,} Mean Absolute Commanded Jerk and Mean Absolute Commanded Acceleration. Mean jerk decreases by 75.6\%, and acceleration decreases by 32.8\%--58.0\% across axes. These metrics characterize command smoothness under the evaluated conditions.}

	\label{fig:xianaopinghua}
\end{figure*}

\begin{figure*}[t]
	\centering
	\includegraphics[width=17cm]{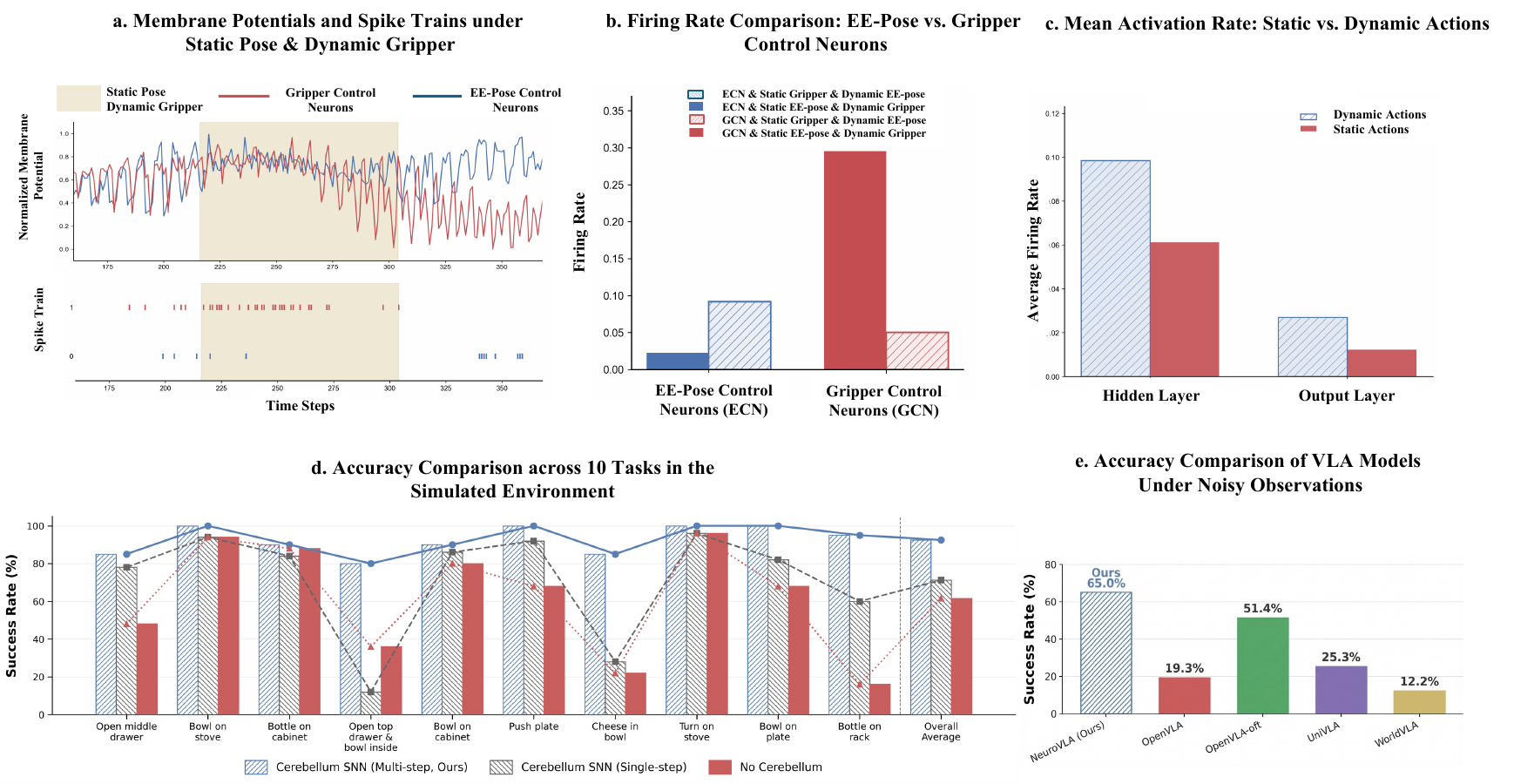}
	\caption{\textbf{Selective spiking activity and temporal ablation.}
\textbf{a,} Membrane potentials and spike trains during ``Static Pose \& Dynamic Gripper'' show relatively quiescent end-effector pose neurons and active gripper-control neurons.
\textbf{b,} Firing rates across end-effector and gripper conditions quantify demand-dependent activity.
\textbf{c,} Mean activation rates are lower in static than dynamic phases.
\textbf{d,} LIBERO task success with a multi-step SNN, a single-step variant, and a no-cerebellum baseline.
\textbf{e,} Task success under noisy observations compares NeuroVLA with the evaluated VLA baselines. Temporal ablation and noisy-observation evaluation test different aspects of the control architecture.}

	\label{fig:snnresult}
    \vspace{-0.4cm}
\end{figure*}

\begin{figure*}[t]
	\centering
	\includegraphics[width=17cm]{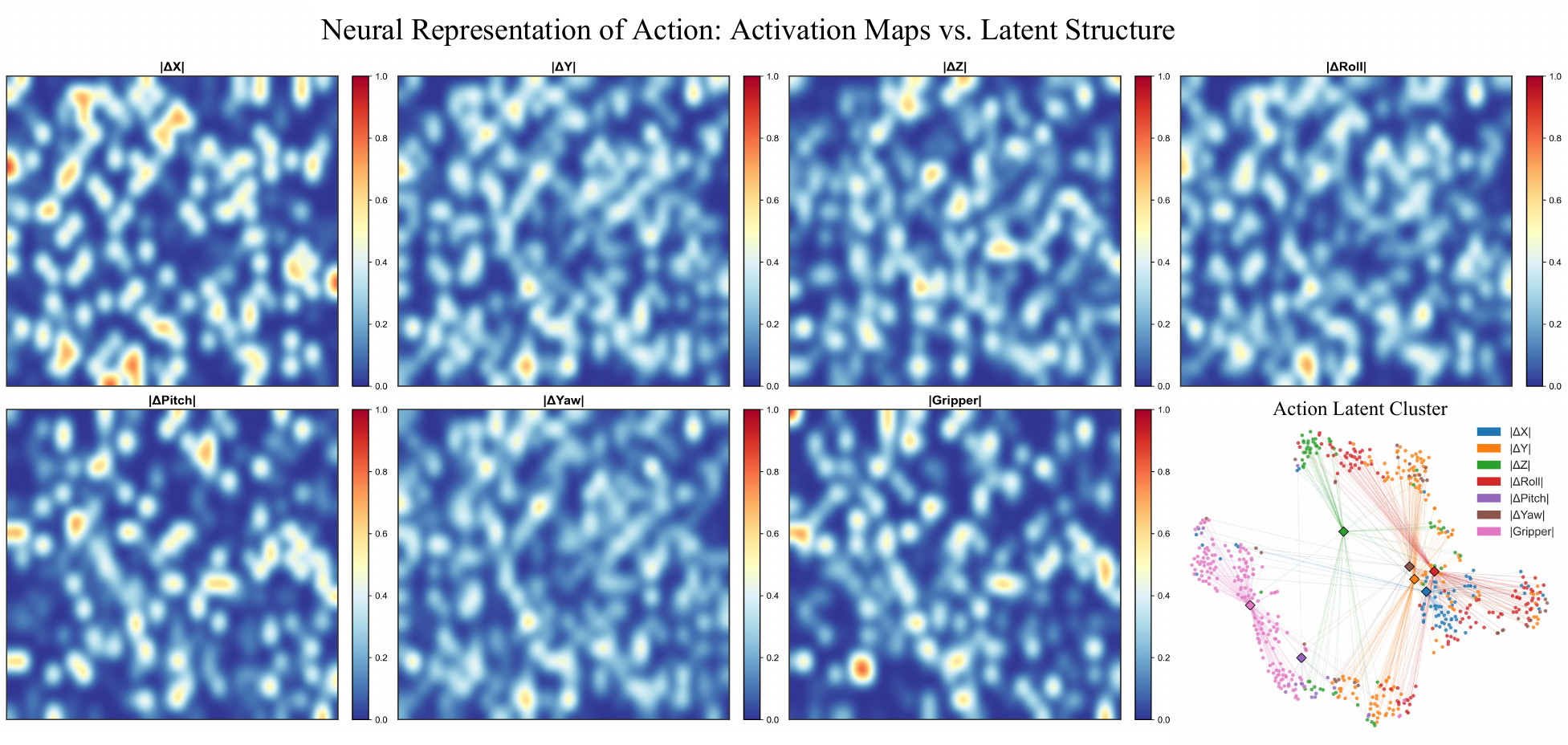}
	\caption{\textbf{Action-associated structure in spinal representations.}
\textbf{Neural Representation of Action:} Rearranged firing-rate maps show different activation patterns for kinematic dimensions such as $|\Delta Roll|$ and $|\text{Gripper}|$.
\textbf{Action Latent Cluster:} A t-SNE projection of spinal hidden states separates several action-associated groups. These visualizations describe action-associated organization in the learned motor representation.}
    \label{fig:relitu}
\end{figure*}

To investigate the mechanistic basis of the system's robustness,
we examined three control functions: stabilization during free-space motion, local correction after contact, and temporal coordination during rhythmic manipulation. These functions motivate the cerebellar design, but the experiments assess robot behaviour rather than identifying separate anatomical loops.

\subsubsection*{Proprioceptive stabilization of ongoing motion}
We first tested whether feedback-based modulation reduces high-frequency fluctuations in commanded motion. In free-space trajectories, the baseline policy produced pronounced acceleration oscillations, whereas activating the cerebellar module attenuated these fluctuations (Fig.~\ref{fig:xianaopinghua}a).

Using end-effector pose, joint angles, and joint velocities as state feedback, the module reduced mean commanded jerk by 75.6\%, with reductions of 80.2\% in $\Delta yaw$ and 80.0\% in $\Delta Z$ (Fig.~\ref{fig:xianaopinghua}b). Mean Absolute Commanded Acceleration (MACA) decreased by 32.8\% to 58.0\% across axes (Fig.~\ref{fig:xianaopinghua}c). These measurements demonstrate smoother command trajectories under the evaluated conditions.

\subsubsection*{Force-aware correction after unexpected contact}
We next tested whether local force feedback enables adaptation when an intended trajectory is obstructed. Contact generated sharp perturbations in the 6-DoF wrench profile (Fig.~\ref{fig:xianaoresult}b), including mean force values of $\mu \approx -37.02$ in $F_x$ and $\mu \approx 8.30$ in $F_y$, with variance $\sigma^2 \approx 3.27$ in $F_y$.

The vision-centric baseline continued along the blocked path (red dashed line). NeuroVLA used the high-frequency ($>200$~Hz) force stream to adjust the local trajectory and navigate around the contact point (blue solid line). This result supports force-aware motor correction through the feedback pathway. The force/torque sensor measures contact state; it is not treated as a substitute for a biological vestibular organ.

\subsubsection*{Temporal coordination during rhythmic manipulation}
In the ``Shake the cup'' task, three oscillatory cycles were visible in synchronized wrench and joint-angle traces (Fig.~\ref{fig:xianaoresult}a). The alignment of these trajectories indicates coordinated rhythmic execution, including periods when visual feedback was static or occluded. These synchronized traces demonstrate coordinated rhythmic execution; the spinal temporal contribution is assessed by the ablation below.

Under degraded visual observations, the cerebellar-enhanced agent also maintained higher success than the evaluated cortical baselines (Fig.~\ref{fig:snnresult}e). This supports the usefulness of proprioceptive context when visual information is unreliable. The temporal contribution of the spinal dynamics is examined separately through the following ablation.

\subsection{Event-driven sparsity and temporal integration in the neuromorphic spinal substrate}
We next examined whether spinal activity follows actuation demand and whether persistent spiking state contributes to task performance. During ``Static Pose \& Dynamic Gripper'' phases, end-effector pose neurons remained relatively quiescent while gripper-control neurons fired more actively (Fig.~\ref{fig:snnresult}a,b). Mean activation rates were lower during static holding than during dynamic actions (Fig.~\ref{fig:snnresult}c). These observations demonstrate selective, demand-dependent neural activity; hardware energy consumption is evaluated separately below.

Stateful spiking neurons carry temporal information by accumulating and decaying membrane voltage~\cite{zhao2025lsnn}. On LIBERO~\cite{liu2023libero, fei2025libero}, the multi-step SNN achieved higher success than a single-step variant, particularly in tasks requiring transitions between manipulation phases (Fig.~\ref{fig:snnresult}d). The same panel includes a no-cerebellum baseline. The temporal ablation supports the contribution of persistent spiking state, while the complementary baseline evaluates the contribution of cerebellar modulation.

Firing-rate maps conditioned on kinematic dimensions showed distinct activation patterns (Fig.~\ref{fig:relitu}, Neural Representation of Action), and t-SNE projections separated several action-associated latent states (Action Latent Cluster). These visualizations suggest functional specialization within the learned controller. They provide descriptive evidence of representation structure rather than demonstrating a biological somatotopic map or establishing the cause of the performance gains.

\subsection{Hardware validation of the spiking spinal controller}
\begin{figure*}[t]
	\centering
	\includegraphics[width=17cm]{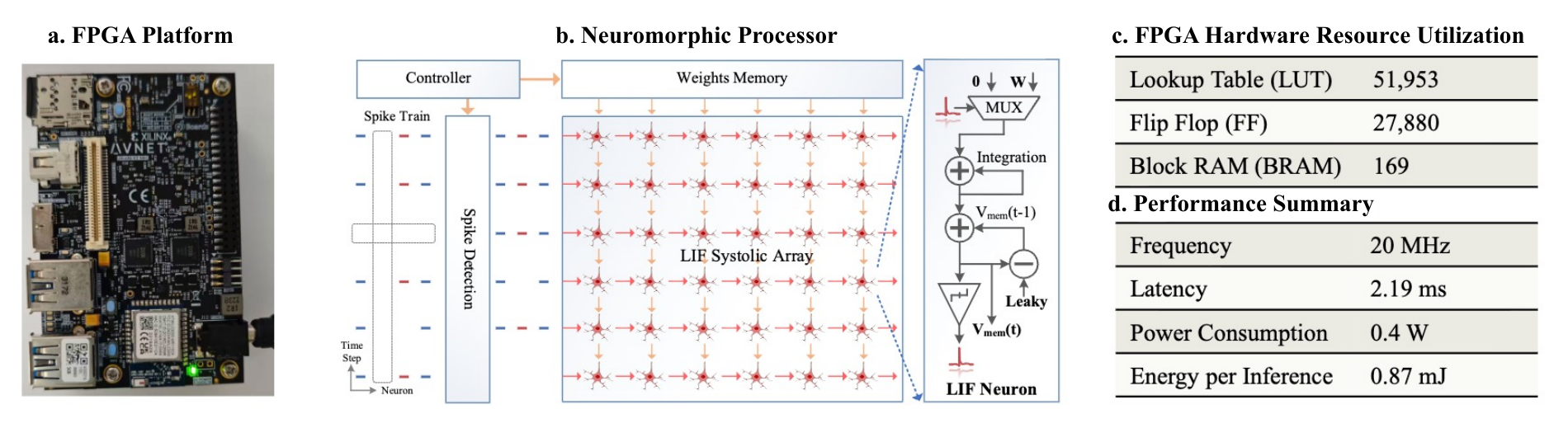}
	\caption{\textbf{Neuromorphic processor implemented on an FPGA platform for deploying the Spiking Spinal Module.}
    \textbf{a,} FPGA board used for implementation. \textbf{b,} Neuromorphic processor incorporating an LIF systolic-array architecture and spike-sparsity-aware computation to reduce inference latency and energy consumption. \textbf{c,} FPGA resource utilization, with 51,953 LUTs, 27,880 FFs and 169 BRAMs. \textbf{d,} Performance summary, showing an inference latency of 2.19 ms at 20 MHz and an energy cost of 0.87 mJ per inference.}
	\label{fig:fpga}
\end{figure*}

We implemented the spinal controller on an FPGA-based neuromorphic processor (Fig.~\ref{fig:fpga}a). Its systolic array updates multiple leaky integrate-and-fire neurons~\cite{abbott1999lapicque} in parallel, reusing spikes across columns and weights across rows (Fig.~\ref{fig:fpga}b). A spike-detection module suppresses inactive events to reduce unnecessary updates~\cite{rao2022long, Rathi2022Exploring}.

At an operating frequency of 20 MHz, the processor achieved 2.19 ms inference latency and 0.87 mJ per inference, corresponding to approximately 0.4 W during inference. Resource utilization and performance are summarized in Fig.~\ref{fig:fpga}c,d. These measurements characterize the spinal controller on the FPGA implementation; they do not include cortical and cerebellar GPU computation, sensing, communication, or robot actuation.

\begin{figure*}[!ht]
	\centering
	\includegraphics[width=17cm]{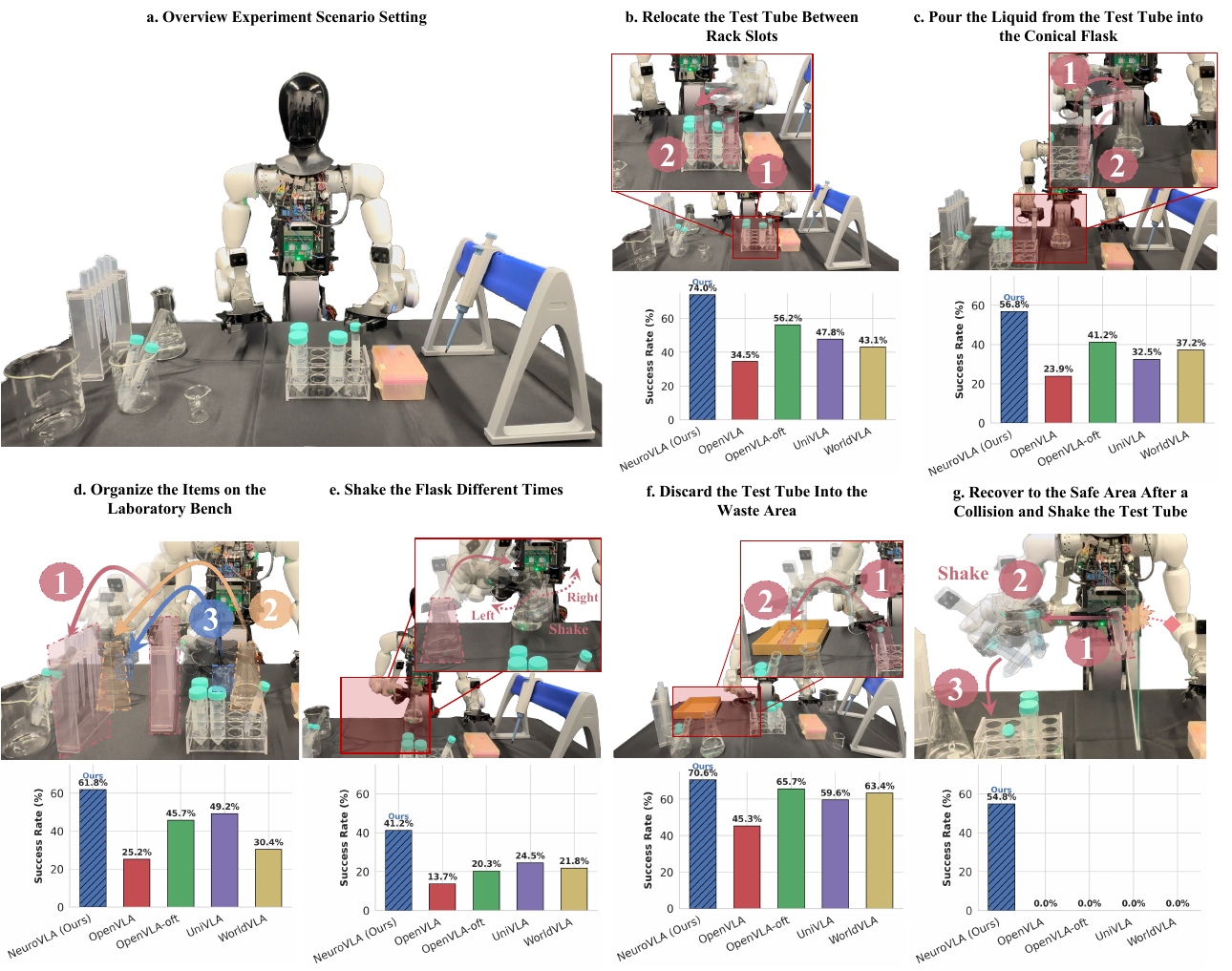}
	\caption{\textbf{Physical manipulation and recovery after unexpected contact.}
\textbf{a,} Bimanual laboratory platform and experimental scene.
\textbf{b--f,} Task success for relocating test tubes (b), pouring liquid (c), organizing laboratory items (d), shaking flasks (e), and discarding test tubes (f), comparing NeuroVLA with OpenVLA, OpenVLA-OFT, UniVLA, and WorldVLA.
\textbf{g,} Recovery after collision followed by test-tube shaking. NeuroVLA achieves 54.8\% recovery compared with 0.0\% for the evaluated baselines. The task tests withdrawal and subsequent continuation under an unexpected physical obstruction.}

	\label{fig:realexresult}
    \vspace{-0.3cm}
\end{figure*}

\subsection{Physical manipulation and collision recovery}
We deployed NeuroVLA on a physical bimanual humanoid platform to test spatial precision, fluid handling, and rhythmic execution (Fig.~\ref{fig:realexresult}a--f). Across the evaluated tasks, it achieved higher success than OpenVLA~\cite{kim2024openvla}, OpenVLA-OFT~\cite{Kim2025FineTuningVM}, UniVLA~\cite{Wang2025UnifiedVM}, and WorldVLA~\cite{Cen2025WorldVLATA}.

During pouring (Fig.~\ref{fig:realexresult}c), changing loads and visual occlusion challenged stable execution. The cerebellar module integrated proprioceptive feedback at 200 Hz, allowing motor commands to be adjusted as the physical state changed. In flask shaking (Fig.~\ref{fig:realexresult}e), coordinated repeated motions were consistent with the temporal context provided by the recurrent and spiking states. These task observations complement the stabilization measurements and temporal ablation above.

We then introduced an unexpected obstruction during the ``Recover to Safe Area'' task (Fig.~\ref{fig:realexresult}g). The evaluated VLA baselines achieved 0.0\% recovery and often continued executing the obstructed trajectory. NeuroVLA achieved 54.8\% recovery. Following a wrench transient, the local cerebellar--spinal pathway generated a withdrawal response with a reported latency below 50 ms, bypassing the cortical inference loop ($>200$ ms). Local trajectory correction then supported recovery and task continuation.

Together with the other physical tasks, the collision result supports the use of bodily feedback and local motor state to complement semantic instruction following. The limits of recovery are considered in the Discussion.

\section{Discussion}
\subsection*{Beyond architectural inertia from language modeling}
The success of language-model pretraining makes its architectures and scaling strategies a natural starting point for embodied intelligence. However, this starting point can become a constraint if motor competence is treated primarily as an extension of semantic sequence generation. Our results motivate a broader design agenda: high-frequency proprioceptive sensing, persistent motor state, and local corrective pathways should be explicit architectural components, rather than capabilities assumed to emerge from semantic scaling alone. This position complements progress in vision-language reasoning and continuous-action experts by addressing the bodily interface through which their outputs are executed.

NeuroVLA addresses the resulting mismatch between semantic inference and physical interaction. The Q-Former supplies task-relevant intent, recurrent cerebellar modulation incorporates bodily feedback, and stateful spinal dynamics generate continuous actions. Attention maps, command-smoothing measurements, temporal ablations, and collision tests provide complementary evidence for these roles. The contribution is an engineered control architecture whose feedback bandwidth and temporal state are matched to motor demands, rather than a demonstration that instruction understanding alone yields stable control.

The cortex, cerebellum, and spinal cord provide functional inspiration for this division of responsibilities~\cite{Ito2008InternalModels, bruel2024spinal}. Our implementation does not reproduce their detailed anatomy or identify a one-to-one mapping to spinocerebellar, vestibulocerebellar, or cerebrocerebellar circuits. Biological correspondence should therefore be interpreted as a design hypothesis, with the engineering claims assessed by the robot experiments.

\subsection*{Motor memory and efficient execution}
The term ``muscle memory'' motivates a concrete design requirement here: a motor controller should retain execution-relevant state and use recent sensorimotor history to shape its next action. In NeuroVLA, recurrent proprioceptive context and persistent membrane dynamics provide this short-term motor memory. The temporal ablation supports the usefulness of retained state for task execution; it does not establish the long-term skill consolidation associated with biological motor learning. Distinguishing these timescales makes the analogy testable and links it to implementable control mechanisms.

The multi-step ablation supports the value of persistent motor state, while selective firing shows how neural activity can follow actuation demand. FPGA measurements establish low latency and energy use for the spinal controller. These findings support hardware-efficient local execution, but low spinal power does not imply that the complete VLA system or robot operates at the same power. The hardware inference measurement is also distinct from end-to-end collision-response latency.

\subsection*{Limits of local recovery and future work}
Local force feedback improved recovery in the tested contact scenario, yet a 54.8\% success rate does not establish a general safety guarantee. Broader perturbation tests and explicit sensing-to-actuation latency measurements would better characterize the operating limits. Comparisons in this study concern the selected baselines and tasks, rather than all VLA architectures.

Training currently relies on offline behaviour cloning and GPU-based surrogate gradients. Although the spinal controller is validated on an FPGA-based neuromorphic processor, the cortical and cerebellar modules remain on the CUDA tier. Online plasticity and deployment of additional control components on neuromorphic hardware are future directions~\cite{pei2019towards}. Simulation studies that independently vary proprioceptive feedback rate, sensory delay, recurrent state, and spiking state would help separate their contributions to stable, temporally coordinated manipulation. Such tests would evaluate the proposed sensorimotor design principles directly, alongside subsequent validation on physical hardware.

\section{Methods}
\label{methods}
\subsection{Architecture Overview}
We formalize the embodied control task as a Partially Observable Markov Decision Process (POMDP)\cite{lauri2022partially}. At each time step $t$, the agent receives a multimodal observation tuple $o_t \in \mathcal{O}$, comprising high-dimensional visual sensory data $I_t \in \mathbb{R}^{H \times W \times 3}$, a natural language instruction $L$, and a history of low-dimensional proprioceptive states $\mathbf{s}_{t-H:t} \in \mathbb{R}^{H \times D_s}$, where $H$ denotes the size of history window and $D_s$ represents the dimensions of joint positions, velocities, and end-effector forces (6-DoF wrench). The objective is to generate a sequence of continuous motor actions $\mathbf{a}_t \in \mathcal{A}$ to maximize the cumulative reward.

To reconcile the conflicting computational requirements of semantic planning (high-latency, abstract) and motor execution (low-latency, precise), we propose the \textbf{Tri-Level Neuromorphic Vision-Language-Action ( NeuroVLA)} architecture. As shown in Fig. 1, the system control policy $\pi(a_t | o_t)$ is decomposed into a hierarchical composition of three specialized mapping functions:$$\mathbf{a}_t = \Phi_{\text{spine}} \left( \Phi_{\text{cerebellum}} \left( \Phi_{\text{cortex}}(I_t, L), \mathbf{h}_t \right) \right),$$ where $\Phi_{\text{cortex}}$, $\Phi_{\text{cerebellum}}$, and $\Phi_{\text{spine}}$ correspond to the cortical, cerebellar, and spinal modules, respectively. $\mathbf{h}_t$ denotes the compact dynamic context vector from the neurological state estimator. This decomposition enforces a strict division of labor across timescales and hardware substrates:

\begin{enumerate}
    \item \textbf{Cortical Semantic Projection ($\Phi_{\text{cortex}}$):} Operates on the CUDA Computing Tier. It maps the visual-linguistic inputs $(I_t, L)$ into a semantic latent intention space $\mathbf{z}_{\text{sem}} \in \mathbb{R}^{D_{\text{model}}}$, where $D_{\text{model}}$ denotes the dimension of model internal feature space. This module utilizes a Vision-Language Model (VLM) backbone to extract open-vocabulary affordances, without requiring a new semantic update at every local control step.

    \item \textbf{Cerebellar State Modulation ($\Phi_{\text{cerebellum}}$):} Operates on the CUDA Computing Tier. It functions as a state-dependent affine transformation. A recurrent estimator aggregates the proprioceptive history $\mathbf{s}_{t-H:t}$ into a dynamic context vector $\mathbf{h}_t$. This context is then used to modulate the semantic latent $\mathbf{z}_{\text{sem}}$ via a Gated Feature-wise Linear Modulation (FiLM) mechanism, yielding a refined, physics-aware latent code $\mathbf{z}_{\text{mod}}$. This process is inspired by cerebellar feedback modulation; it conditions the action latent on observed physical state.

    \item \textbf{Spinal Neuromorphic Decoding ($\Phi_{\text{spine}}$):} Operates on Neuromorphic Chip Tier. It maps the modulated latent $\mathbf{z}_{\text{mod}}$ to continuous actions $\mathbf{a}_t$ via a Spiking Neural Network (SNN) governed by Leaky Integrate-and-Fire (LIF) dynamics. This module leverages event-driven sparsity and intrinsic temporal memory to ensure energy-efficient, smooth trajectory generation and rapid reflex responses.
\end{enumerate}

The entire system is trained end-to-end using a hybrid loss that combines behavior cloning losses with surrogate gradients\cite{neftci2019surrogate} for the non-differentiable spiking components, ensuring functional coherence across the hierarchy.

\subsection{Cortical Module: Semantic Latent Generation}
\label{sec:cortical_module}

The Cortical Module occupies the highest tier of the control hierarchy, tasked with cross-modal reasoning and the synthesis of abstract motor intentions from high-dimensional observations. Unlike lower-level circuits that must react to instantaneous physical dynamics, the cortical layer operates on a semantic timescale, prioritizing long-horizon planning and generalization over high-frequency control.

\subsubsection{Vision-Language Reasoning Backbone}
We leverage a pre-trained Large Vision-Language Model (VLM), specifically Qwen-VL\cite{bai2023qwenvl}, as the general-purpose reasoning engine. Formally, given the current RGB observation $I_t$ and the natural language instruction $L$, the VLM functions as a dense feature extractor:$$\mathcal{H}_t = F_{\text{VLM}}(I_t, L; \theta_{\text{vlm}}),$$
where $\mathcal{H}_t = \{h_t^{(1)}, \dots, h_t^{(N)}\}$ represents the stack of hidden states across all $N$ transformer layers and $\theta_{\text{vlm}}$ denotes the parameters of VLM. This rich representation captures scene geometry cues, object semantics, and linguistic context, providing a multimodal scene representation that supports open-vocabulary reasoning and manipulation.

\subsubsection{Layer-wise Semantic Distillation}
Directly feeding the high-dimensional, variable-length VLM hidden states into a motor policy is computationally prohibitive and prone to overfitting. To bridge the dimensionality gap between the VLM's linguistic-visual space and the robot's control manifold, we employ a \textbf{Layer-wise Querying Transformer (Q-Former)}.

The Q-Former acts as a learnable information bottleneck. It utilizes a set of learnable query tokens $\mathbf{Q} \in \mathbb{R}^{K \times D}$ to attend to specific subsets of the VLM's hidden representations, where $K$ denotes the number of tokens and $D$ denotes the token dimension. Crucially, rather than utilizing only the final layer, we extract features from a designated range of intermediate layer index $[l_{\text{start}}, l_{\text{end}}]$ to capture both low-level spatial details and high-level semantic abstractions. The extraction process is formalized as:$$\mathbf{z}_{\text{sem}} = \text{Q-Former}(\text{Concat}(\mathcal{H}_t[l_{\text{start}}:l_{\text{end}}]), \mathbf{Q}; \theta_{\text{Q-Former}}),$$ where $\mathbf{z}_{\text{sem}} \in \mathbb{R}^{K \times D_{\text{action}}}$ denotes the \textbf{Semantic Latent Intention} and $\theta_{\text{Q-Former}}$ denotes the parameters of Q-Former.

\textbf{Design Interpretation:} The Q-Former compresses image-language features into the semantic latent $\mathbf{z}_{\text{sem}}$, providing task intent to the downstream controller. Physical context is incorporated through the separate cerebellar modulation described below.

\subsection{Cerebellar Module: Gated Recurrent Neuromodulation}
\label{sec:cerebellar_module}

While the cortical module generates semantic intentions, it operates significantly slower than the dynamics of physical interaction and lacks access to high-frequency proprioceptive feedback. To bridge this gap, the Cerebellar Module functions as an adaptive filter, performing \textit{dynamic state estimation} and \textit{trajectory refinement} before the signals reach the spinal execution layer.

\subsubsection{Proprioceptive State Estimation via Recurrent Dynamics}
\label{sec:GRU}
Inspired by the integration of bodily feedback in biological motor control, we employ a Gated Recurrent Unit (GRU)\cite{chung2014empirical} as a recurrent state estimator.
Let $\mathbf{s}_{t-H:t} \in \mathbb{R}^{H \times D_s}$ denote the history of robot states over a horizon $H$ (aligned with the sensor frequency, e.g., 50 Hz), where each state vector includes joint angles, velocities, and 6-DoF wrench measurements. The estimator integrates this temporal sequence to produce a compact dynamic context vector $\mathbf{h}_t$:$$\mathbf{h}_t = \text{GRU}(\mathbf{s}_{t-H:t}; \theta_{\text{gru}}),$$
where $\theta_{\text{gru}}$ denotes the parameter of the designed GRU. Unlike static encoders (e.g., MLPs), the GRU captures the \textit{rate of change} and \textit{contact transients} (e.g., collision impulses), which are critical for stabilizing motion under perturbation.

\subsubsection{Gated Feature-wise Linear Modulation (FiLM)}
\label{sec:FiLM}
The core computational role of the cerebellum is to modulate cortical commands based on current physical constraints. We formalize this interaction using a Gated FiLM mechanism\cite{perez2018film}.
First, to prevent proprioceptive noise from overwhelming semantic intent during stable phases, we employ a learnable gating factor $\mathbf{g}_t$:
$$\mathbf{g}_t = \sigma(W_g \cdot \text{Proj}(\mathbf{h}_t)),$$ where $\sigma$, $W_g$ and $\text{Proj}(\cdot)$ denote the sigmoid function, the learnable affine transformation matrix and the projection function. This gate selectively regulates how much physical context is allowed to influence the cortical plan.
Subsequently, the dynamic context $\mathbf{h}_t$ is projected into affine transformation parameters---scale $\boldsymbol{\gamma}_t$ and shift $\boldsymbol{\beta}_t$---which are applied to the cortical semantic latent $\mathbf{z}_{\text{sem}}$:
$$\boldsymbol{\gamma}_t = f_{\gamma}(\mathbf{h}_t), \quad \boldsymbol{\beta}_t = f_{\beta}(\mathbf{h}_t)$$ $$\mathbf{z}_{\text{mod}} = (1 + \boldsymbol{\gamma}_t) \odot (\mathbf{z}_{\text{sem}} \cdot \mathbf{g}_t) + \boldsymbol{\beta}_t$$, where $\odot$ denotes the element-wise Hadamard product. This operation allows the cerebellar module to strictly enforce gain control: for instance, upon detecting a collision (spike in $\mathbf{h}_t$), the network can suppress the forward velocity encoded in $\mathbf{z}_{\text{sem}}$ (via $\boldsymbol{\gamma}_t \approx -1$) and inject a retraction bias (via $\boldsymbol{\beta}_t$), effectively rewriting the motor plan in real-time.

\subsubsection{Iterative Refinement of the Motor Latent}
The motor latent is updated through an \textit{Iterative Refinement Loop}. Given an updated anticipated state, the module $\text{Refine}(\cdot)$ re-applies the two cerebellar computations introduced in Sections \ref{sec:GRU} and \ref{sec:FiLM} to adjust the motor latent. The module performs $K$ cycles of internal recurrence (set to $K=2$ in our experiments).
In each iteration $k \in \{0,...,K-1\}$, the module predicts a tentative action latent, updates the anticipated state evolution, and re-modulates the input. Formally:$$\mathbf{z}_{\text{mod}}^{(k+1)} \leftarrow \text{Refine}(\mathbf{z}_{\text{mod}}^{(k)}, \mathbf{s}_{t+1})$$
The recurrence allows the motor latent to be revised before spinal action decoding. An explicit decomposition into identified gravity, friction, or other physical-model terms is not imposed by this formulation.

\textbf{Design Interpretation:} This architecture uses the functional separation of intended action and bodily feedback. The semantic latent $\mathbf{z}_{\text{sem}}$ provides a descending intention signal, while $\mathbf{h}_t$ summarizes proprioceptive history. FiLM conditions the motor latent on this context. This design is inspired by cerebellar feedback and internal-model accounts~\cite{Ito2008InternalModels}; the formulation does not by itself demonstrate an explicit sensory-prediction-error computation or anatomical equivalence.

\subsection{Spinal Module: Spiking Residual Dynamics}
\label{sec:spinal_module}

The Spinal Module serves as the final execution interface, translating the modulated latent intentions $\mathbf{z}_{\text{mod}}$ into continuous motor commands $\mathbf{a}_t$. Implemented on neuromorphic principles, this module is designed to support energy-efficient event-driven processing and continuous action decoding via intrinsic neuronal dynamics.

\subsubsection{Stateful Leaky Integrate-and-Fire Dynamics}
We model the spinal interneurons using the iterative Leaky Integrate-and-Fire (LIF) formalism. Crucially, the network is instantiated with \textbf{stateful membrane dynamics}, meaning the membrane potential $u_i^{(l)}$ is strictly preserved across successive time steps rather than being re-initialized at each forward pass. This endows the spinal substrate with an \textit{implicit temporal working memory}, allowing it to encode history-dependent features without requiring explicit recurrent gating units (e.g., LSTMs). For a hidden neuron $i$ in layer $l$ at time step $\tau$, the dynamics evolve according to:$$u_i^{(l)}[\tau] = \beta u_i^{(l)}[\tau-1] + \sum_j w_{ij} s_j^{(l-1)}[\tau] - s_i^{(l)}[\tau-1] \cdot \vartheta,$$ where $\beta \in (0, 1)$ is the membrane decay factor, $w_{ij}$ is the synaptic weight, $s_j^{(l-1)}[\tau] \in \{0, 1\}$ represents the incoming spike train and $\vartheta$ denotes the voltage. The term $u_i^{(l)}[\tau-1]$ represents the \textit{residual potential} carried over from the previous moment, establishing a continuous temporal context.

\subsubsection{Deep Spiking Residual Architecture}
To enable the learning of complex sensorimotor transformations without signal degradation in deep spiking layers, we structure the spinal module as a Spiking ResNet\cite{fang2021deep, hu2021spiking}. By incorporating residual skip connections:$$\mathbf{x}^{(l+1)} = \mathbf{x}^{(l)} + \text{LIF}(\text{Linear}(\mathbf{x}^{(l)})).$$

We facilitate the direct propagation of gradients and spike rates during training. Residual connections facilitate the propagation of the modulated latent through the spiking layers.

\subsubsection{Continuous Integration for Motor Decoding}
A critical challenge in applying SNNs to robotics is bridging the gap between discrete, stochastic spikes and the continuous domain of joint actuation. We resolve this by implementing a \textbf{Continuous Integration Protocol} at the output layer\cite{guo2024spgesture}. Unlike hidden interneurons which undergo rapid repolarization (reset) upon spiking, the output motor neurons are configured as \textbf{non-resetting integrators}. They continuously accumulate synaptic influx over the simulation window without discharging their potential:$$\mathbf{a}_t[\tau] = \mathcal{W}_{\text{out}} \cdot \mathbf{u}_{\text{out}}[\tau],$$
where $\mathbf{u}_{\text{out}}$ represents the accumulated membrane voltage. This mechanism serves two control functions:
\begin{enumerate}
    \item \textbf{Temporal Smoothing:} The integration process acts as a natural low-pass filter, converting high-frequency spike trains into smooth, continuous motor trajectories. This mirrors the biological process of \textit{twitch summation}, where muscles integrate neural drive into continuous force, filtering out neural noise.
    \item \textbf{Kinematic Consistency:} By maintaining the voltage state, the network naturally enforces temporal consistency in the output action sequence, reducing kinematic jerk and ensuring fluid transitions between control steps.
\end{enumerate}

\subsubsection{Surrogate Gradient Learning}
To overcome the non-differentiability of the spike generation function $s = \Theta(u - \vartheta)$, we employ a surrogate gradient method for end-to-end training. During the backward pass, the Heaviside step function is approximated by a fast sigmoid function $\sigma(x) = \frac{x}{1 + |x|}$, allowing gradients to flow from the continuous action space back through the discrete spinal layers to the cerebellar and cortical modules.

\section*{Data Availability}
The simulation datasets used in this study are derived from the open-source \textbf{LIBERO} and \textbf{LIBERO-Plus} benchmarks. The real-world experimental data collected during the current study are available from the corresponding author upon reasonable request.

\section*{Code Availability}
The code of this paper is available at \url{https://github.com/guoweiyu/NeuroVLA}.

\section*{Acknowledgements}
This work was supported in part by the National Key R \& D Program of China (Grant No.2023YFF0725001), in part by the National Natural Science Foundation of China (Grant No.92370204), in part by the Guangdong Basic and Applied Basic Research Foundation (Grant No.2023B1515120057), in part by the Education Bureau of Guangzhou.
\section*{Author Contributions}
W.Y.G. proposed the idea of NeuroVLA,  implemented the core codebase, and analyzed the results.
P.T.L  designed the real-world experiments, test the performance and discuss the idea.
Z.H,T.F.C and Z.Y.C designed the explainable experiments,analyzed the results and discuss the idea.
Y.S, Y.K.Y and H.X supervised the literature review, data processing, methodology, analysis, and discussion.
Y.D.G, X.H and H.X managed this project.

\section*{Competing Interests}
The authors declare no competing interests.

\section*{Inclusion \& Ethics Statement}
We confirm that this study adheres to all applicable ethical regulations.

\end{document}